\documentclass[10pt,twocolumn,letterpaper]{article}

\usepackage[pagenumbers]{wacv} % To force page numbers, e.g. for an arXiv version

\usepackage{xpinyin}
\usepackage{enumitem}
\newcommand{\chinese}[1]{%
  \begin{CJK}{UTF8}{gbsn}#1\end{CJK}%
}
\usepackage{microtype}
\usepackage{graphicx}
\usepackage{amsmath}
\usepackage{multirow}
\usepackage{booktabs}
\usepackage{easyeqn}
\usepackage{bm}
\usepackage{setspace}
\usepackage{float}
\usepackage{amsfonts}
\usepackage{amssymb}
\usepackage{textcomp}
\usepackage{listings}
\usepackage[table]{xcolor}
\usepackage{colortbl} 
\usepackage[most]{tcolorbox}
\lstdefinestyle{plain}{
    basicstyle=\fontsize{9}{10}\ttfamily,
    keywordstyle=\color{blue},
    commentstyle=\color{gray},
    stringstyle=\color{green},
    showstringspaces=false,
    breaklines=true,
    breakatwhitespace=false,
    breakindent=0pt,
    escapeinside={(*@}{@*)}
}

\usepackage{multicol}
\usepackage{multirow}
\usepackage{CJK}

\definecolor{wacvblue}{rgb}{0.21,0.49,0.74}
\usepackage[pagebackref,breaklinks,colorlinks,allcolors=wacvblue]{hyperref}

\def\wacvPaperID{3293} % *** Enter the WACV Paper ID here
\def\confName{WACV}
\def\confYear{2026}

\title{Vision Language Models Are Not (Yet) Spelling Correctors }

\author{
Junhong Liang$^{1,}$\thanks{Equal contribution} \quad 
Bojun Zhang$^{2,}$\footnotemark[1] \\
$^{1}$MBZUAI \quad $^{2}$Institute of Automation, Chinese Academy of Sciences \\
{\tt\small junhong.liang@mbzuai.ac.ae, zhangbojun2022@ia.ac.cn}
}
\begin{document}
\maketitle
\begin{abstract}
% The ABSTRACT is to be in fully justified italicized text, at the top of the left-hand column, below the author and affiliation information.
% Use the word ``Abstract'' as the title, in 12-point Times, boldface type, centered relative to the column, initially capitalized.
% The abstract is to be in 10-point, single-spaced type.
% Leave two blank lines after the Abstract, then begin the main text.
% Look at previous \confName\ abstracts to get a feel for style and length.

Spelling correction from visual input poses unique challenges for vision–language models (VLMs), as it requires not only detecting but also correcting textual errors directly within images. We present ReViCo (Real Visual Correction), the first benchmark that systematically evaluates VLMs on real-world visual spelling correction across Chinese and English. ReViCo contains naturally occurring errors collected from real-world image data and supports fine-grained evaluation at both image and token levels. Through comprehensive experiments on representative cascaded (Qwen) and native (InternVL) open-source models, as well as closed-source systems (GPT-4o, Claude), we show that current VLMs fall significantly short of human performance, particularly in correction. To address these limitations, we explore two solution paradigms: a Joint OCR–Correction pipeline and a Background Information enhanced approach, both of which yield consistent performance gains. Our analysis highlights fundamental limitations of existing architectures and provides actionable insights for advancing multimodal spelling correction.

% Spelling correction from visual input poses challenges beyond conventional text-only methods, as models must both detect and correct errors directly from images. We introduce ReViCo, a novel benchmark for \textbf{R}eal \textbf{Vi}sual \textbf{Co}rrection task in Chinese and English, covering both erroneous and error-free samples. A wide range of vision–language models (VLMs) are evaluated, including cascaded architectures (Qwen2.5-VL), native multimodal designs (InternVL3.5), and closed-source systems (GPT-4o, Gemini, Claude). Our results show that detection is consistently easier than correction; for example, QwenVL-32B achieves an $F_{1}$ score of 64.3 on image-level detection but only 25.5 on image-level correction. We also find that smaller InternVL models outperform larger QwenVL variants, highlighting the impact of architecture over scale. Finally, we explore a Joint-OCR correction paradigm that improves performance by providing explicit textual cues. These findings underscore the limitations of current VLMs and suggest that future progress depends on both architectural advances and more effective task formulations.
\end{abstract}
  
\section{Introduction}
\label{sec:introduction}

% Spelling correction plays a pivotal role in refining textual data by rectifying erroneous characters or words. While traditional approaches have focused on text-based error correction, the emergence of spelling errors in vision-based scenarios necessitates specialized solutions. Addressing these errors is critical for ensuring the accuracy and efficacy of subsequent NLP tasks. 

Vision--language models (VLMs) have recently demonstrated remarkable progress on multimodal tasks such as visual question answering (VQA)~\cite{jian-etal-2024-large}, spatial reasoning~\cite{zhang2025mllmsstrugglespatialunderstanding}, and optical character recognition (OCR)~\cite{nagaonkar2025benchmarkingvisionlanguagemodelsoptical}. These tasks primarily test a model’s ability to understand or transcribe visual input. However, an important capability remains underexplored: \emph{visual error correction}---the task of detecting and correcting textual errors directly from images.  

\begin{figure}
    \centering
    \includegraphics[width=\linewidth]{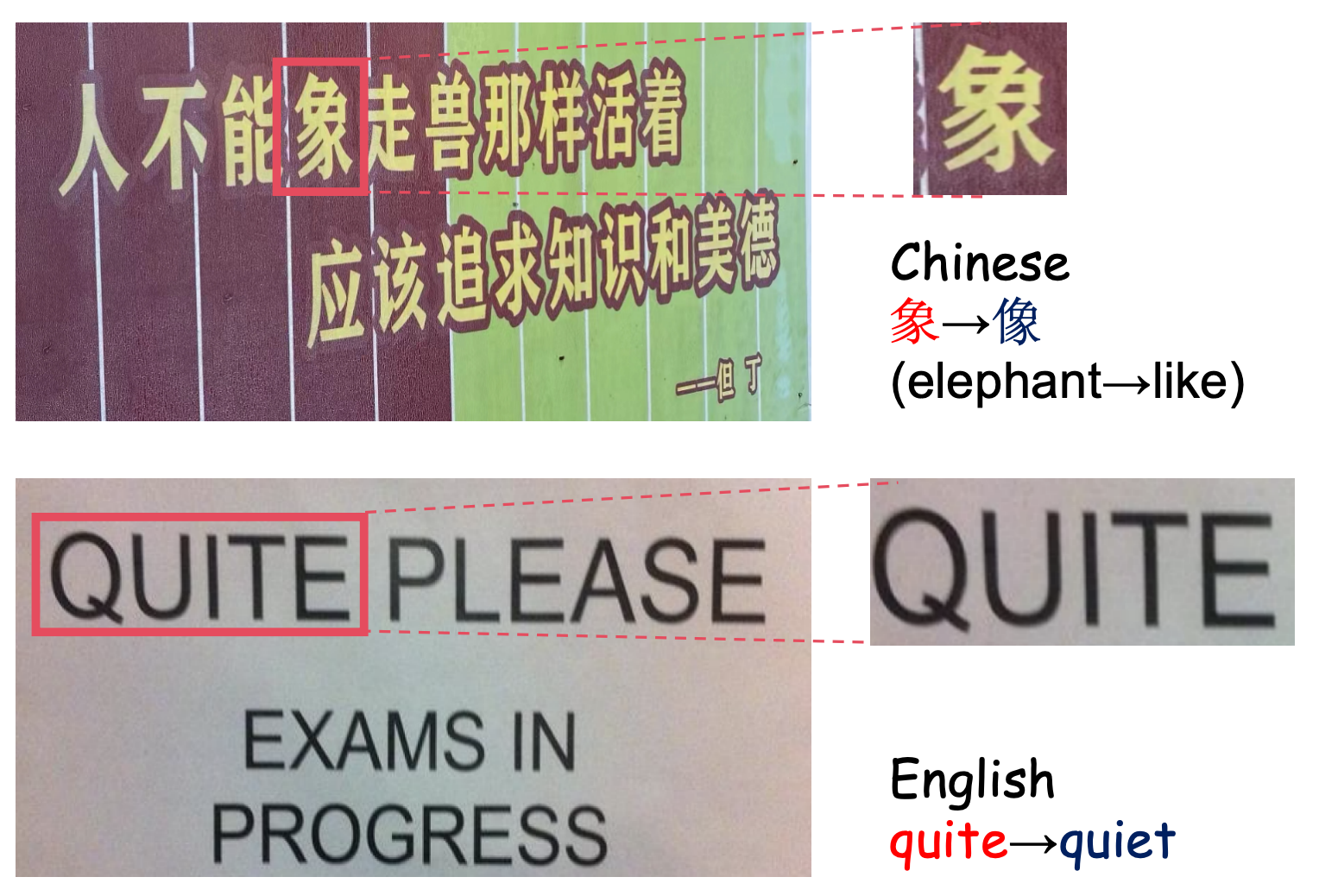}
    \caption{Examples of visual error correction in Chinese (\chinese{象} $\rightarrow$ \chinese{像}) and English (``quite'' $\rightarrow$ ``quiet'')}
    \label{fig:examples_intro}
\end{figure}

Most existing studies treat error correction as an OCR post-processing problem, where the OCR output is assumed to be accurate and correction is formulated as a purely text-centric language modeling or sequence-to-sequence task. Yet OCR errors are common, and they can propagate to downstream applications such as document image translation~\cite{liang-etal-2024-document}, question answering~\cite{wang2020generalvalueevidencebilingual}, and table structure recognition~\cite{10.1007/978-981-97-8511-7_27}. This limitation motivates a broader view: correcting errors not only at the text level, but by grounding the correction in the \emph{visual context} of the image itself.  Figure~\ref{fig:examples_intro} demonstrates examples in Chinese and English, where the errors are created by mistyping of similar characters/words.

Based on the analysis above, we propose a fundamental research question: \emph{Can VLM correct visual spelling errors?} We divide this question into two subtasks: (1) \emph{detecting} which tokens are incorrect based on both textual and visual cues, and (2) \emph{restoring} them to their correct form. Unlike OCR post-processing, this requires VLMs to jointly leverage fine-grained visual perception and linguistic reasoning within a unified inference process.

To address this issue, we construct ReViCo, the first benchmark for real-world visual error correction. ReViCo features of high-quality Chinese and English images that contain natural spelling errors, captured in everyday scenarios. Using this benchmark, we systematically conduct the first comprehensive evaluation of VLMs on visual error correction, including both open-source and closed-source models. For the open-source models, we use two main types of VLMs: the cascaded structure Qwen and the native multimodel design InternVL. For closed-source models, we use GPT-4o and Claude-sonnet. We discover that even for the best models, there exists a gap between human evaluation and the performance of VLMs.

To address the limitations, we further propose two solution paradigms: a Joint OCR--Correction approach, which integrates explicit text recognition with correction, and a Background Information Enhanced approach, which exploits contextual visual cues. Finally, we provide an in-depth analysis of current VLM limitations, offering insights for future research on multimodal error correction.  

Our main contributions are as follows:  
\begin{enumerate}
    \item We firstly define the vision spelling correction task and release ReViCo, a novel dataset for visual spelling correction in Chinese and English, where all data are collected from real scenarios.  
    \item We benchmark a wide range of VLMs and propose solutions Joint OCR--Correction and Background Information Enhanced approach) for improved performance.  
    \item We present detailed findings on the strengths and weaknesses of current VLMs, establishing a foundation for future advances in multimodal correction.  
\end{enumerate}

\section{Related Works}
\label{sec:related_works}

\paragraph{Architecture of VLMs}
The architectures of VLMs can be broadly categorized into two paradigms: \emph{cascaded} and \emph{native multimodal} designs. Cascaded architectures, such as Qwen2.5-VL~\cite{bai2025qwen25vltechnicalreport}, employ a pretrained vision encoder (e.g., CLIP or ViT) to transform images into visual tokens, which are then projected into the embedding space of an LLM. This lightweight alignment layer enables reuse of strong unimodal backbones, but may introduce an information bottleneck that hinders fine-grained transfer of perceptual details. In contrast, native multimodal models such as InternVL~\cite{wang2025internvl35advancingopensourcemultimodal} adopt an integrated backbone where both image and text tokens are jointly processed within a unified transformer. By eliminating the projection bottleneck, this design achieves tighter cross-modal alignment and often exhibits superior performance, particularly in reasoning-intensive or fine-grained visual tasks. Recent studies highlight that architectural choices, rather than model scale alone, critically determine the effectiveness of VLMs on challenging benchmarks.

Beyond open-source efforts, several proprietary systems have explored hybrid or advanced integration strategies. GPT-4o~\cite{openai2024gpt4technicalreport}, Gemini, and Claude exemplify closed-source models that combine multimodal encoders with large-scale language backbones, often with additional optimizations for long-context reasoning, safety alignment, and robustness~\cite{li2025surveystateartlarge}. Although details are not fully disclosed, their performance provides an upper bound for comparison with open-source systems.

Overall, these architectural divergences, specifically the cascaded projection versus native multimodal fusion, form the foundation for analyzing VLM performance in our benchmark. As our results later show, structural design often outweighs parameter count in determining success on vision spelling correction tasks.

% \textbf{Using Language Models for Spelling Error Correction}
% Spelling Error Correction has been a long-established task, aiming at detect and correct the wrong spelled token in a sentence. Current Spelling Error Correction includes XX and XX in Chinese, XX and XX

% Current spelling correction systems could be divided into two categories, Pretrained Language MOdels (PLM) and Large Language Models(LLM), in PLM, people use pretrained Bert scale language models for spelling correction, such as XX and XX. LLMs have demonstrated superior performance in a series of tasks, could also be applied for spelling correction, relevant research includes XX and XX. 

\paragraph{Scenarios for Spelling Error Correction}
Spelling Error Correction is a long-standing task that aims to \emph{detect} and \emph{correct} misspelled tokens within sentences. In Chinese, common challenges arise from \emph{phonetic similarity} (homophones or near-homophones at the pinyin level) and \emph{graphical similarity} (visually confusing characters) \cite{cheng-etal-2020-spellgcn,liu-etal-2021-plome,zhang-etal-2020-spelling}, while in English it's related to letter changes in a word\cite{electronics9101670,mitton1980birkbeck}.
Existing datasets are primarily text-based, including those collected from language learners~\cite{tseng-etal-2015-introduction, dahlmeier-etal-2013-building}, native speakers~\cite{hu-etal-2024-cscd}, and specific domains~\cite{liang2025rairretrievalaugmentediterativerefinement,jiang2022mcscset,lv2023general}. Other scenarios have also been explored, such as Automatic Speech Recognition (ASR)~\cite{liang2024perlpinyinenhancedrephrasing, Ma_2023} and student essay writing~\cite{li-etal-2024-towards-real}.

% In English, errors are commonly divided into \emph{nonword errors} (out-of-vocabulary strings) and \emph{real-word errors} (valid words used in contextually inappropriate ways, e.g., ``desert'' vs.\ ``dessert'') \cite{electronics9101670}. Several spelling error corpora have been investigated: the Birkbeck corpus examines errors made by adult native speakers, while Holbrook analyzes errors produced by children in English-speaking countries\footnote{\url{https://titan.dcs.bbk.ac.uk/~ROGER/corpora}}. In addition, Wikipedia provides a curated list of common misspellings collected from its editors\footnote{\url{https://en.wikipedia.org/wiki/Wikipedia:Lists_of_common_misspellings}}. 

\begin{figure*}[t!]
  \includegraphics[width=\linewidth]{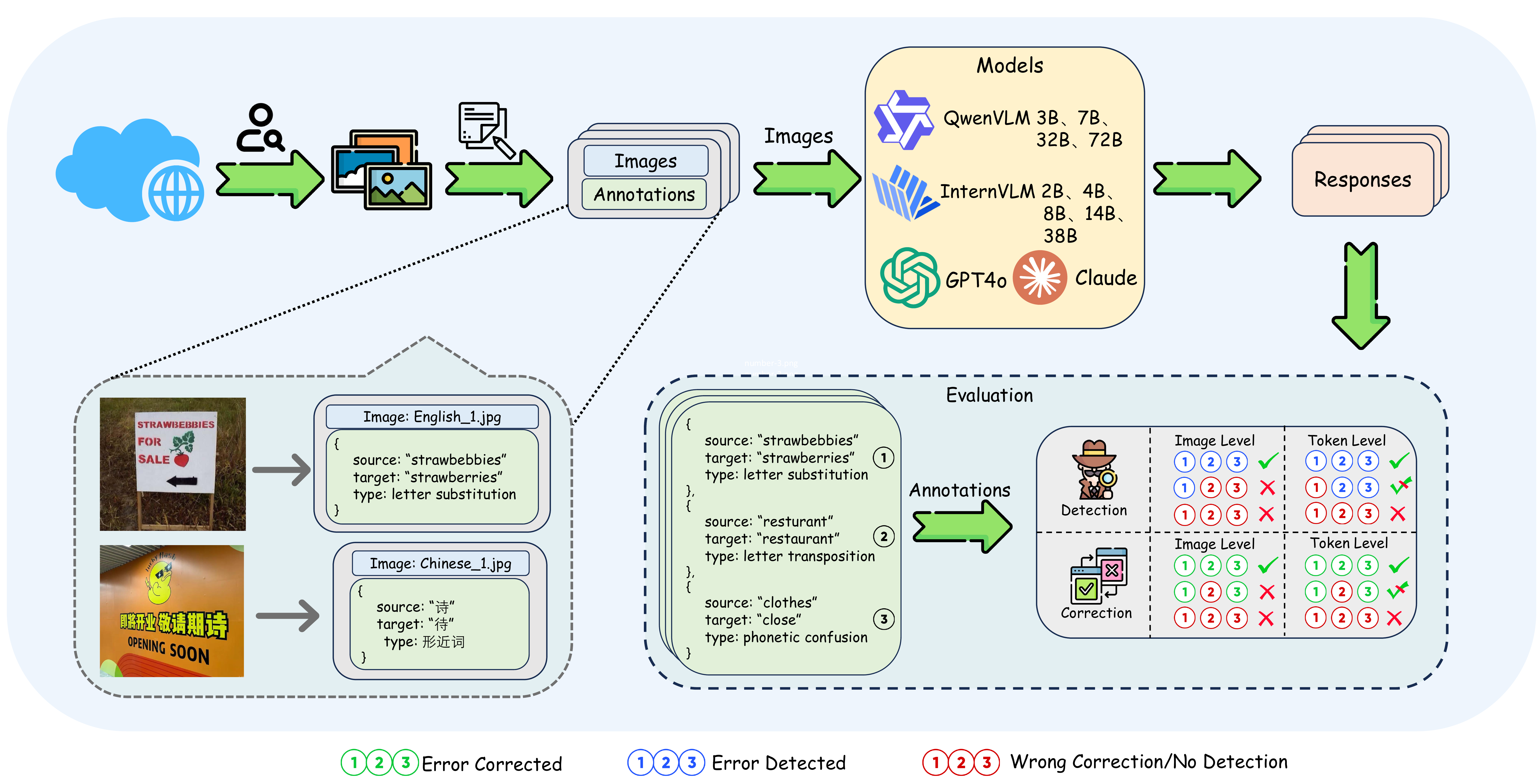}
  \caption{Overview of the ReViCo Benchmark pipeline, illustrated with one English and one Chinese example from the dataset. The evaluation procedure is shown with a single English example.}
  \label{fig:framework}
  \vspace{-4mm}
\end{figure*}

\paragraph{PLM vs.\ LLM}
Contemporary approaches to text spelling correction can be broadly divided into two paradigms: Pretrained Language Models (PLM) and Large Language Models (LLM).  
(i) \textbf{PLM-based methods:} These approaches leverage pretrained encoder models such as BERT. For instance, Soft-Masked BERT explicitly models error positions through a soft masking mechanism and performs correction jointly \cite{zhang-etal-2020-spelling}. SpellGCN incorporates phonological and visual similarities via graph convolution networks to improve character-level correction \cite{cheng-etal-2020-spellgcn}. PLOME further enhances pretraining by integrating confusion sets and multi-granularity phonological/orthographic knowledge \cite{liu-etal-2021-plome}, while SoMu integrates Wubi decomposition features, which lie between character-level and stroke-level representations, and simultaneously investigates both Chinese spelling correction and splitting correction~\cite{liang-etal-2024-hybrid}. 
(ii) \textbf{LLM-based methods:} More recently, LLMs have shown strong potential for spelling correction without extensive task-specific tuning. For example, \citet{zhou-etal-2025-training} demonstrates that LLMs can function as effective language model decoders for spelling correction in a training-free and prompt-free setting. Other work has proposed hybrid approaches that dynamically combine the probabilities of smaller PLMs with LLMs during decoding to balance domain precision with generalization \cite{qiao2025mixturesmalllargemodels}. Comprehensive surveys further highlight the transition from PLM to LLM paradigms and identify open challenges and opportunities in this domain  \cite{liu2025chinesespellingcorrectioncomprehensive}.

\section{ReViCo Benchmark}
\label{sec:visualcsc_benchmark}
% Bojun did statistics of the dataset
To systematically analyze the visual error correction ability of LLMs, we propose \textbf{Re}al \textbf{Vi}sual \textbf{Co}rrection (ReViCo) Benchmark. As most of the current spelling correction benchmarks focus on the text modality, there is a need to bridge the gap between modalities. In this section, we introduce our ReViCo Benchmark. Section \ref{subsec:task_design} presents the task design, Section \ref{subsec:construction_pipeline} shows the construction pipeline of our processed data, Section \ref{subsec:data_statistics} presents the statistics of our proposed data. An overview of the data gathering and evaluation process is shown in Figure~\ref{fig:framework}.

\begin{figure*}
    \centering
    \includegraphics[width=\linewidth]{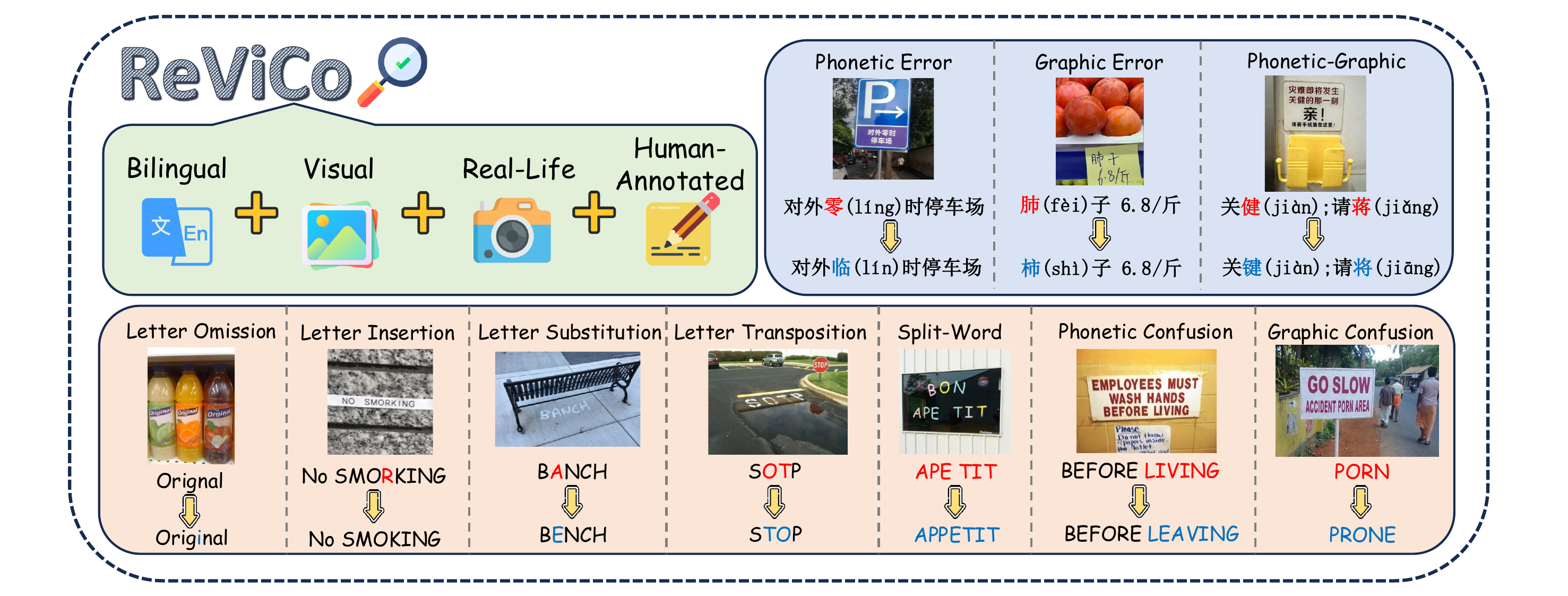}
    \caption{Examples from our ReViCo Benchmark, where the error types for Chinese and English are displayed. Wrong tokens are highlighted in \textcolor{red}{red} and the corresponding correct tokens are highlighted in \textcolor{blue}{blue}.}
    \label{fig:error_type_examples}
    \vspace{-1em}
\end{figure*}
\subsection{Task Design}
\label{subsec:task_design}

To evaluate the error correction ability of VLMs, we design two tasks: \emph{Visual Spelling Error Detection} and \emph{Visual Spelling Error Correction}.

\textbf{Visual Spelling Error Detection} requires the model to determine whether an image contains spelling errors, specifically erroneous characters in Chinese or misspelled words in English. Given an input image, the model must decide whether the image has spelling errors.

\textbf{Visual Spelling Error Correction} requires the model to not only detect but also correct the spelling errors in the image. To achieve this, the model must accurately identify the erroneous tokens within the text and replace them with their correct forms.

\subsection{Construction Pipeline}
\label{subsec:construction_pipeline}

To construct the visual spelling error correction dataset, we collected images from multiple online resources. Three volunteers were invited to participate in the annotation process. They manually searched for Chinese and English images containing spelling errors through a web-based collection. To comprehensively evaluate the error detection ability of VLMs, we additionally sampled error-free Chinese and English images from EST-VQA dataset ~\cite{wang2020generalvalueevidencebilingual}  .

The collected error-containing images were randomly divided into two subsets. Each subset was annotated by one volunteer and subsequently cross-checked by another, with the task of identifying misspelled tokens and providing the corresponding correct forms. In cases of disagreement between annotators, the third volunteer served as an adjudicator to determine the final label. Through this multi-stage annotation and verification process, we obtained a high-quality labeled dataset for visual spelling error correction.

\subsection{Data Statistics}
\label{subsec:data_statistics}

\textbf{Real error types} ReViCo Benchmark consists of datasets in two languages, Chinese and English, which contain four and seven subsets, respectively. The Chinese dataset can be divided into two broader categories: \textit{phonetic error} (i.e., substitutions involving characters with similar pronunciation), \textit{graphic error} (i.e., substitutions involving characters with similar visual appearance), \textit{phonetic-graphic error} (i.e., cases that simultaneously involve both phonetic and graphic similarity), and others(i.e., instances that do not fall into the preceding categories). 

Due to the structural differences in the minimal linguistic units, errors in the English dataset can be classified at two distinct levels of granularity: the letter level and the word level. At the letter level, errors can be further classified into four fine-grained categories: \textit{letter omission} (i.e., a letter is missing from a word), \textit{letter insertion} (i.e., an extra letter is incorrectly added to a word), \textit{letter substitution} (i.e., a correct letter is replaced by an incorrect one), \textit{letter transposition} (i.e., two adjacent letters are swapped). At the word level, errors can be divided into two broader categories, generally corresponding to \textit{split-word errors} (i.e., a single word is incorrectly broken into two or more), \textit{phonetic/graphic confusion} (i.e., a word is incorrectly used in place of another with a similar sound or spelling), and others.

This hierarchical categorization provides a systematic framework for analyzing error types and contributes to a more precise evaluation of text correction methods. Examples of different types of errors are presented in Figure~\ref{fig:error_type_examples}. \\

% We also provide examples of our proposed dataset, the Chinese and English error samples are presented in Table \ref{tab:example_Chinese} and Table \ref{tab:example_English}, respectively.
% \input{table/example_Chinese}
% \input{table/example_English}

\begin{figure}
    \centering
    \includegraphics[width=\linewidth]{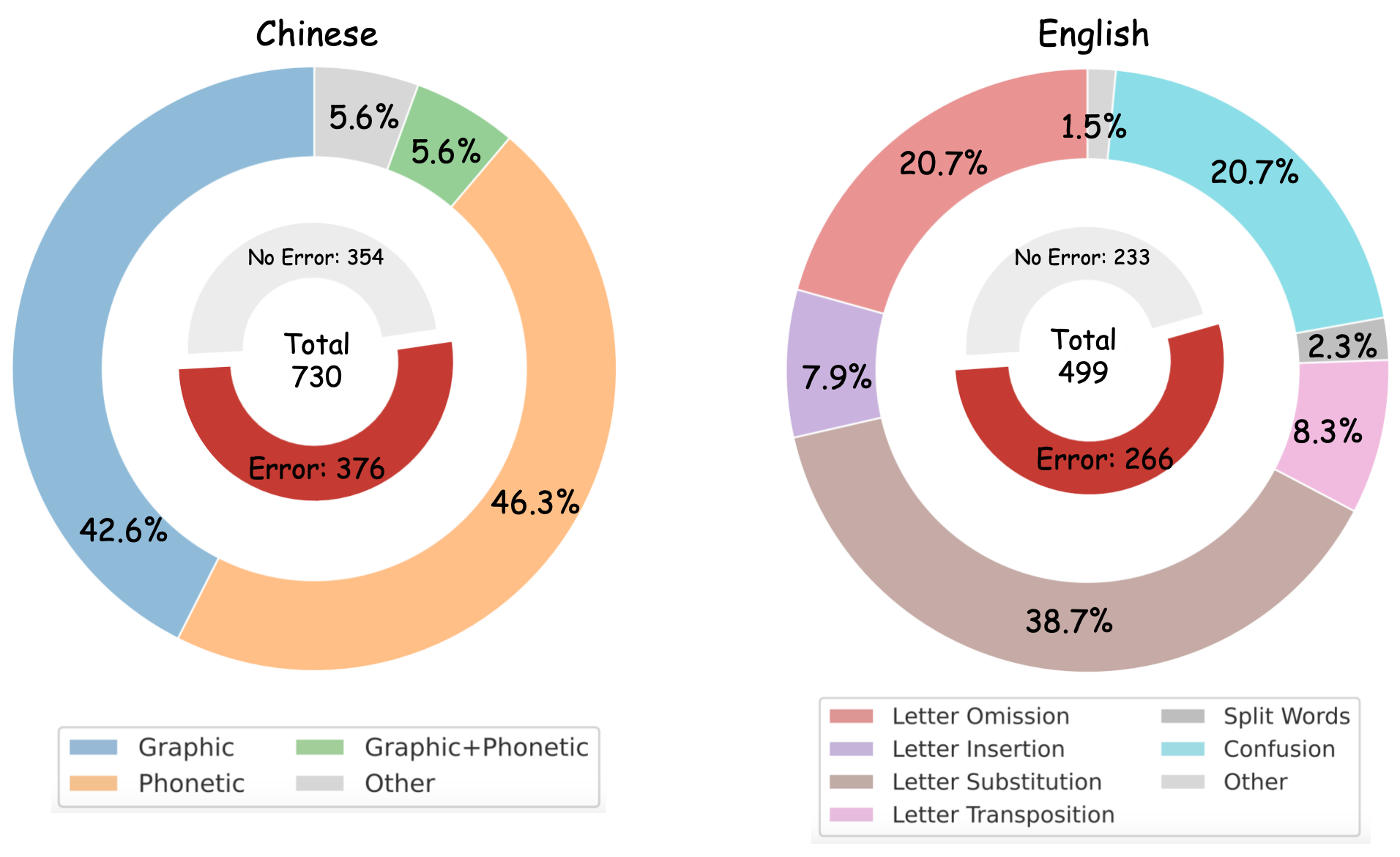}
    \caption{Error statistics of the ReViCo benchmark. For Chinese and English separately, the inner pie charts show the counts of erroneous vs. error-free images, while the outer pie charts depict the distribution of error types across the erroneous images.}
    \label{fig:error_statistics_revico}
    \vspace{-1.5em}
\end{figure}

\textbf{Error Type Analysis} Figure~\ref{fig:error_statistics_revico} presents the statistics of error types. To assess the error detection ability of VLMs, we also include error-free images as part of the evaluation. For Chinese data, phonetic and graphic errors dominate, reflecting common scenarios of character misspellings in printing or handwriting. In English, the most frequent error type is letter substitution, which typically arises from similar-looking letters or adjacent key presses during typing.

\begin{table*}[t!]
\centering
\caption{Performance comparison of VLMs and humans on spelling error detection and correction, evaluated at both the image (I) level and token (T) level across detection (D) and correction (C) tasks.  The best overall results in each column are highlighted in \textbf{bold}, while the strongest results within the QwenVL and InternVL families are \underline{underlined}.}
\label{tab:baseline_result_qwen_intern_gpt_new}
\scalebox{1.1}{
\begin{tabular}{lcccccc|cccccc}
\toprule
%  & \multicolumn{6}{c}{\textbf{Detection}} & \multicolumn{6}{c}{\textbf{Correction}} \\ 
% \cmidrule(lr){2-7} \cmidrule(lr){8-13}
\multirow{2}{*}{Models} & \multicolumn{3}{c}{\textbf{I}mage-\textbf{D}etection} & \multicolumn{3}{c}{\textbf{T}oken-\textbf{D}etection} 
 & \multicolumn{3}{c}{\textbf{I}mage-\textbf{C}orrection} & \multicolumn{3}{c}{\textbf{T}oken-\textbf{C}orrection} \\ 
\cmidrule(lr){2-4} \cmidrule(lr){5-7} \cmidrule(lr){8-10} \cmidrule(lr){11-13}
  & $P$ & $R$ & $F_1$ & $P$ & $R$& $F_1$ & $P$ & $R$ & $F_1$ & $P$ & $R$ & $F_1$ \\
\midrule
QwenVL$_{3B}$
& 48.7 & \underline{73.3} & 58.5  &  5.6 & 16.6 &  8.3 &  6.4 &  5.2 &  5.7  &  5.1 &  7.1 &  6.0 \\ 
QwenVL$_{7B}$ 
 & 70.9 & 25.2 & 37.1  & 28.7 & 12.8 & 17.7 & 45.0 &  8.4 & 14.2 & 29.1 &  9.9 & 14.7 \\ 
QwenVL$_{32B}$ 
 & \underline{96.9} & 48.1 & 64.3  & 58.2 & \underline{31.8} & \underline{41.1} & \underline{94.1} & \underline{25.0} & \underline{39.5} & \underline{52.8} & \underline{27.9} & \underline{36.5} \\ 
QwenVL$_{72B}$
 & 93.8 & 49.4 & \underline{64.7} & \underline{59.7} & 31.2 & 41.0 & 87.9 & 23.8 & 37.5 & 50.7 & 24.5 & 33.1 \\
 \hline 
InternVL$_{2B}$ 
 & 53.8 & 40.1 & 45.9 &  5.8 & 11.7 &  7.8 &  5.8 &  2.1 &  3.1 &  3.3 &  4.0 &  3.7 \\ 
InternVL$_{4B}$
 & 55.8 & \underline{67.7} & \underline{61.2} &  5.8 & 14.5 &  8.3 &  8.7 &  5.1 &  6.5 &  5.0 &  6.2 &  5.5 \\ 
InternVL$_{8B}$
 & 76.7 & 21.2 & 33.2 & 12.8 &  8.1 &  9.9 & 34.9 &  3.5 &  6.3 & 10.5 &  4.9 &  6.7 \\
InternVL$_{14B}$
 & 65.1 & 53.0 & 58.4 & 13.6 & 17.4 & 15.3 & 27.4 & 10.7 & 15.4 & 16.3 & 12.6 & 14.2 \\ 
InternVL$_{38B}$ 
 & \underline{89.8} & 33.4 & 48.7 & \underline{48.6} & \underline{18.9} & \underline{27.2} & \underline{79.3} & \underline{14.5} & \underline{24.5} & \underline{46.5} & \underline{16.1} & \underline{23.9} \\
\midrule
GPT\textendash4o   
& \textbf{98.7} 	& 48.1 	& 64.7  	& \textbf{66.2} 	& \textbf{33.0} 	& \textbf{44.01} 	& \textbf{97.7} 	& \textbf{26.4} 	& \textbf{41.6} 	& \textbf{59.0} 	& \textbf{30.0} 	& \textbf{38.8} \\
Claude\textendash Sonnet\textendash4   
& 70.0 & \textbf{67.8} & \textbf{68.8} & 27.1 & 27.7 & 27.4 & 43.3 & 22.2 & 29.4 & 34.4 & 24.9 & 28.9 \\
\bottomrule
\end{tabular}
}
\end{table*}

\subsection{Evaluation Protocol}
\label{subsec:evaluation_protocol}
To evaluate the capability of correcting visual spelling errors, including both Chinese characters and Latin letters, we define two evaluation tasks, detection and correction. For each task, there are two levels: image and token.

\textbf{Detection} focuses on the capacity of the model to recognize the presence of spelling mistakes. At the \textit{Image Detection} (I-D), the task is to decide whether an image contains any errors at all, treating the problem as a binary classification. In contrast, the \textit{Token Detection} (T-D) moves to a finer granularity, requiring the system to accurately pinpoint the individual erroneous Chinese characters or English words within error-correction pairs. 

\textbf{Correction} evaluates the model's ability to produce correct outputs once errors have been identified. The \textit{Token Correction} (T-C) measures whether the model can transform each erroneous character or word into its proper form, emphasizing precision at the token level. The \textit{Image Correction} (I-C), by comparison, demands a holistic solution: the image is considered successfully corrected only when all errors are resolved without omission or unnecessary alterations. 
%This strict criterion makes it the most challenging setting, as it penalizes both missed corrections and over-corrections.

We report the precision ($P$), recall ($R$) and $F_1$ scores for each task and for each level.

\section{Benchmark Results}\label{sec:benchmark_results}

\subsection{Experiment Setup}

% We evaluate a diverse set of vision-language models (VLMs) as baselines, covering both open-source and closed-source paradigms as classified by \cite{zhang2025mllmsstrugglespatialunderstanding}.
% Among open-source cascaded architectures, we adopt \textbf{Qwen2.5-VL} \cite{qwen2025qwen25technicalreport} and \textbf{LLaVA-OneVision} \cite{li2024llavaonevisioneasyvisualtask}, which connect a pretrained vision encoder (e.g., ViT, CLIP) with a large language model via a lightweight projection layer. 
% We also include the \textbf{InternVL3.5} family \cite{wang2025internvl35advancingopensourcemultimodal}, representing native (monolithic) models where visual perception is tightly integrated into the language backbone. 
% For broader comparison, we consider several influential closed-source systems: \textbf{GPT-4o} \cite{openai2024gpt4technicalreport}, which integrates multimodal reasoning capabilities into OpenAI’s flagship model; \textbf{Gemini} (Google DeepMind), which emphasizes long-context multimodal reasoning; and \textbf{Claude 3.5} (Anthropic), a proprietary model reported to exhibit strong spatial reasoning and safety alignment. 
% Together, these baselines span different architectural paradigms (cascaded vs.~native) and accessibility settings (open vs.~closed source), providing a comprehensive benchmark for evaluating spatial understanding.
\textbf{Model and Baseline} We evaluate a diverse set of open-source vision--language models (VLMs) as baselines. 
Within cascaded architectures, we adopt \textbf{Qwen2.5-VL} \cite{qwen2025qwen25technicalreport}, 
which connects a pretrained vision encoder to a large language model through a lightweight projection layer. 
In contrast, we include the \textbf{InternVL3.5} \cite{wang2025internvl35advancingopensourcemultimodal}, 
which exemplifies native (monolithic) designs where visual perception is tightly integrated into the language backbone. 
For Qwen2.5-VL, we consider models of size 7B, 32B, and 72B, while for InternVL3.5, 
we adopt 2B, 4B, 8B, 14B and 38B variants. 
These selections span multiple architectural paradigms and model scales, 
providing a broad and systematic basis for comparison across cascaded and native approaches.

For closed-source baselines, we consider two widely adopted proprietary models. \textbf{GPT-4o} \cite{openai2024gpt4technicalreport} integrates multimodal reasoning into OpenAI's flagship model, and \textbf{Claude-Sonnet-4} (Anthropic) has been reported to achieve strong performance on spatial reasoning while prioritizing safety alignment. These closed-source systems complement the open-source VLMs, offering a comprehensive benchmark for evaluating spatial understanding across different accessibility settings.  We deploy open-source VLMs on two NVIDIA A100 GPUs, using the default settings of each model. For closed-sourced VLMs, we use official APIs \footnote{The GPT-4o model is gpt-4o-2024-1120 and Claude is claude-sonnet-4-20250514}. The prompt for VLMs is shown in Table \ref{tab:prompt_visual_spelling}.

\begin{table}[t]
\centering
\caption{Prompt templates for English and Chinese Visual Spelling Checking.}
\label{tab:prompt_visual_spelling}
\begin{tabular}{p{0.45\linewidth} p{0.45\linewidth}}
\hline
\multicolumn{2}{l}{\textbf{English Visual Spelling Checking}} \\
\hline
\multicolumn{2}{p{0.9\linewidth}}{
You are an English spelling correction tool. Please check the spelling of the text in the image and these rules exactly:

1. Check every word and if errors exist, output in this exact format: [(''misspelled1'', ''correct1''), (''misspeled2'', ''correct2''), ...], and if no errors, output: [''no error''];

2. Only check English words' spelling.
} \\
\hline
\multicolumn{2}{l}{\textbf{Chinese Visual Spelling Checking}} \\
\hline
\multicolumn{2}{p{0.9\linewidth}}{
\chinese{你是一个严格遵循指令的中文拼写检查工具。请你检查图片中的中文拼写错误，请执行以下操作:  
1. 检查所有汉字，若存在拼写错误，则按照下面的格式输出错误的字及其对应的正确字:[("错误字","正确字")，("错别字","正确字"), ...]；若未发现拼写错误，则输出：["no error"];  

2. 请只关注中文拼写的错误。
}
} \\
\hline
\end{tabular}
\end{table}

\subsection{Result Analysis}

\paragraph{Benchmark Results}
Our benchmark results are presented in Table \ref{tab:baseline_result_qwen_intern_gpt_new}, where we report the performance of open-source VLMs across detection and correction tasks. In our evaluation, \textit{detection} requires the model to identify whether an image contains an error at the image level, and whether incorrect characters or words are detected at the token level. By contrast, \textit{correction} requires the VLMs not only to recognize errors but also to generate the corrected token or the corrected image-level text. Consequently, the requirement for image detection (I-D) is the lowest, whereas image correction (I-C) represents a more challenging task. Across different models, performance is consistently highest in I-D, where Claude achieves 68.8 and QwenVL-72B achieves an $F_{1}$ score of 64.7. However, for I-C, the $F_{1}$ score drops to 41.6 and 25.5, respectively, highlighting that producing corrected output is considerably more complex than binary error detection.  Similarly, QwenVL-32B obtains an $F_{1}$ score of 41.1 on token detection (T-D). Still, its performance decreases to 36.5 on token correction (T-C), a 4.7\% drop, suggesting that while models can often identify erroneous tokens, they struggle to produce the correct replacements. 

\textbf{Open-source VLM Comparison.} We observe a notable phenomenon, that both for QwenVL and InternVL families, small models such as QwenVL-3B and InternVL-4B achieve the best performance in I-D recall (73.3 and 67.7), where we found the models tend to assume the image as erroneous, increasing recall while may over-correct error-free images, therefore diminishing Precision. For larger models, the model tends to assume that the images are error-free, achieving high precision scores for QwenVL-32B (96.9) and InternVL-38B (89.8).

% having fewer parameters. We attribute this to structural differences rather than scale. Qwen2.5-VL uses a cascaded design that projects visual features into the language space, limiting fine-grained perception. In contrast, InternVL employs a native multimodal backbone that jointly models images and text within one transformer, achieving stronger alignment and more efficient parameter use. 

\textbf{Scaling} Surprisingly, increasing the size of QwenVL does not guarantee improved performance. In fact, QwenVL-72B shows a 3\% drop in token-level correction (T-C) compared to QwenVL-32B (36.5 vs. 33.1). Under our direct prompting setting, where models are instructed to output results in a fixed format, we observed that QwenVL-32B often generates additional reasoning steps even though such reasoning was not explicitly requested. This unintended behavior appears to provide auxiliary guidance, leading to higher correction accuracy. In contrast, QwenVL-72B adheres more strictly to the given instructions, producing outputs without supplementary reasoning, which may limit its performance.

\begin{figure}[t!]
    \centering
    \includegraphics[width=\linewidth]{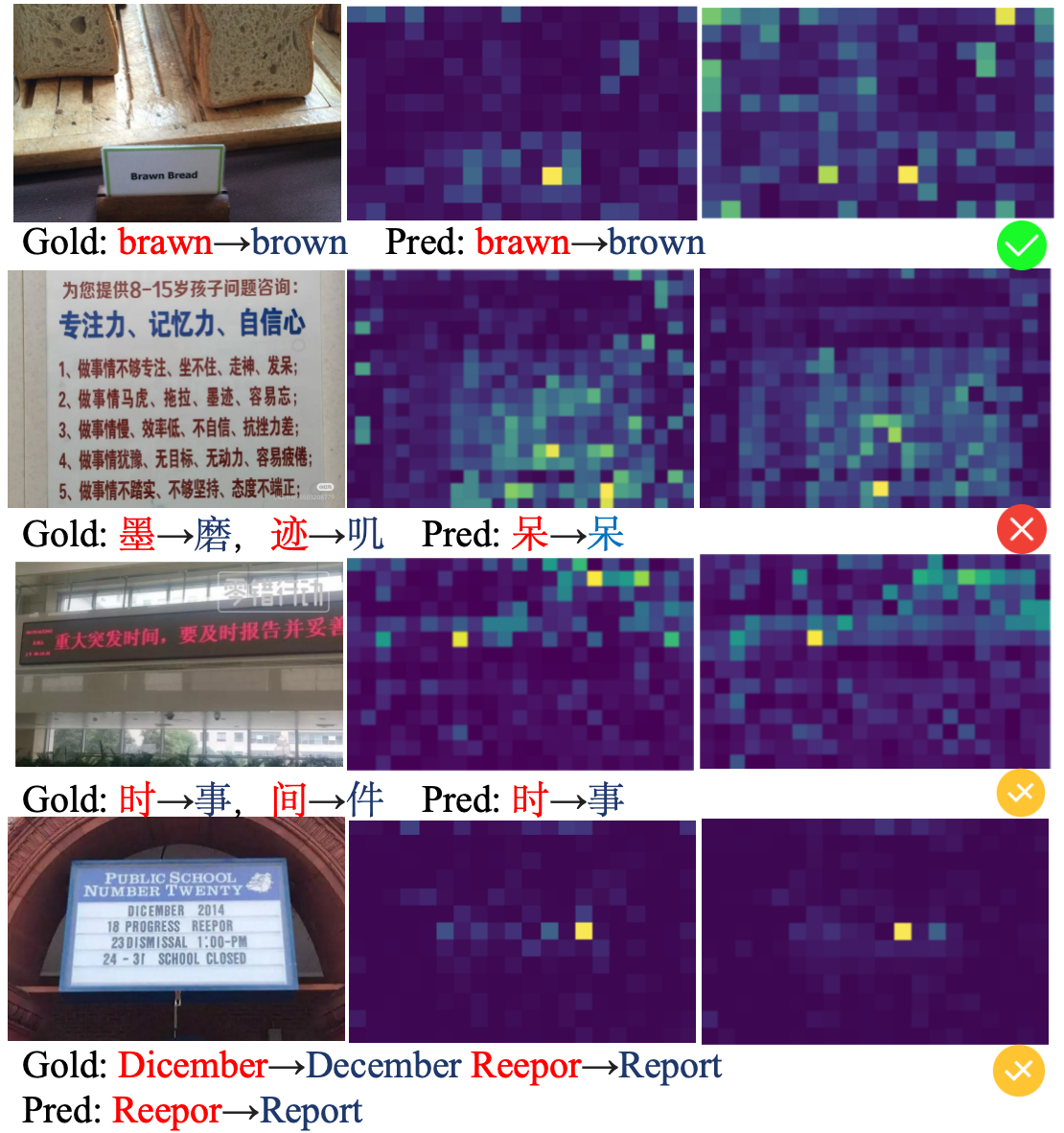}
    \caption{Case studies of attention visualization with QwenVL-7B.Each row shows the original image alongside attention maps from the 22nd and 25th layers. The corresponding golden labels and model predictions are displayed below, where incorrect tokens are highlighted in \textcolor{red}{red} and corrected tokens in \textcolor{blue}{blue}.}
    \label{fig:case_study}
    \vspace{-2em}
\end{figure}

\textbf{Closed-source Model Performance.} For recent proprietary VLMs, such as GPT-4o, performance is generally stronger than most open-source models, achieving 41.3 and 38.6 in I-C and T-C, while Claude-Sonnet-4 achieves the best score in I-D (68.8). Demonstrating GPT-4o has the strongest correction ability in all our models. Nevertheless, some open-source systems (e.g., QwenVL-32B) achieve results comparable to GPT-4o (36.5 vs. 38.6 in T-C), despite the latter having significantly more parameters.

% \paragraph{Instruct Following Ability}

% We summarize the sentences that  could not follow the instruction based on different model settings. 

% \paragraph{Effect of model size}

% We summarize the effect of model size on each question

\paragraph{\textit{How do VLMs recognize error tokens in an image?}}\label{para:attention_visualization_analysis} A case study is illustrated in Figure~\ref{fig:case_study}. The attention mechanism of QwenVL-7B successfully identifies the error in Case 1, demonstrating its ability to detect certain types of mistakes. In Case 2, however, the scenario is more complex: the model fails to attend to the erroneous characters and consequently produces an incorrect output. Notably, it even predicts the character "\chinese{呆}" which does not appear in the image, indicating that under challenging conditions with a large number of tokens, the model is prone to hallucination. Cases 3 and 4 further reveal that when multiple errors are present in the same image, the model can correct one instance while overlooking others, suggesting that its attention mechanism struggles to capture multiple error locations simultaneously. We employ the methodology proposed by \citet{zhang2025mllmsknowlooktrainingfree} to visualize the attention maps.

\begin{table*}[ht]
\centering
\caption{Comparison of F1 scores between the joint OCR-correction paradigm and the direct paradigm. We report results for Detection (D) and Correction (C) at both Image (I) and Token (T) levels. $F_{1d}$ denotes the direct prompt, while $F_{1o}$ represents the joint OCR paradigm (Section \ref{subsec:joint ocr_correction} and $F_{1b}$ represents the joint background information paradigm (Section \ref{subsec:joint_background_information}). Differences are highlighted in \textcolor{red}{red} for decreases and \textcolor{blue}{blue} for increases, all with respect to $F_{1d}$}
\label{tab:joint_ocr_f1_comparision}
\setlength{\tabcolsep}{3pt}
\begin{tabular}{lccc|ccc|ccc|ccc}
\toprule
\multirow{3}{*}{\textbf{Model}} & \multicolumn{3}{c}{\textbf{I-D}} & \multicolumn{3}{c}{\textbf{I-C}} & \multicolumn{3}{c}{\textbf{T-D}} & \multicolumn{3}{c}{\textbf{T-C}} \\
 & ${F_{1d}}$ & ${F_{1o}}$ & ${F_{1b}}$ & ${F_{1d}}$ & ${F_{1o}}$ & ${F_{1b}}$ & ${F_{1d}}$ & ${F_{1o}}$ & ${F_{1b}}$ & ${F_{1d}}$ & ${F_{1o}}$ & ${F_{1b}}$ \\
\midrule
QwenVL$_{3B}$   & 58.5 & 23.6$_{\textcolor{red}{-34.9}}$ & 51.1$_{\textcolor{red}{-7.4}}$ & 8.3  & 5.0$_{\textcolor{red}{-3.3}}$& 0.6$_{\textcolor{red}{-7.7}}$  & 5.7  & 0.6$_{\textcolor{red}{-5.1}}$ & 2.8$_{\textcolor{red}{-2.9}}$ & 6.0  & 1.2$_{\textcolor{red}{-4.8}}$ & 1.5$_{\textcolor{red}{-4.5}}$ \\
QwenVL$_{7B}$   & 37.1 & 60.6$_{\textcolor{blue}{+23.5}}$ & 62.0$_{\textcolor{blue}{+24.9}}$ & 17.7 & 26.7$_{\textcolor{blue}{+9.0}}$ & 20.4$_{\textcolor{blue}{+2.7}}$ &14.2 & 22.7$_{\textcolor{blue}{+8.5}}$ & 21.6$_{\textcolor{blue}{+7.4}}$ & 14.7 & 20.3$_{\textcolor{blue}{+5.6}}$ & 18.5$_{\textcolor{blue}{+3.8}}$ \\
QwenVL$_{32B}$  & 64.3 & 68.9$_{\textcolor{blue}{+4.6}}$ & 66.5$_{\textcolor{blue}{+2.2}}$ & 41.1 & 45.4$_{\textcolor{blue}{+4.3}}$ & 44.5$_{\textcolor{blue}{+3.4}}$ & 39.5 & 42.8$_{\textcolor{blue}{+3.3}}$ & 47.2$_{\textcolor{blue}{+7.7}}$ & 36.5 & 44.0$_{\textcolor{blue}{+7.5}}$ & 43.6$_{\textcolor{blue}{+7.1}}$\\
QwenVL$_{72B}$  & 64.7 & 68.6$_{\textcolor{blue}{+3.9}}$    & 67.8$_{\textcolor{blue}{+3.1}}$          & 41.0 & 45.3$_{\textcolor{blue}{+4.3}}$      & 46.6$_{\textcolor{blue}{+5.6}}$      & 37.5 & 46.8$_{\textcolor{blue}{+9.3}}$     &48.6$_{\textcolor{blue}{+11.1}}$       & 33.1 & 45.0$_{\textcolor{blue}{+11.9}}$    &45.8$_{\textcolor{blue}{+12.7}}$        \\
\hline 
InternVL$_{2B}$ & 45.9 & 35.9$_{\textcolor{red}{-10.0}}$ & 40.8$_{\textcolor{red}{-5.1}}$ & 7.8  & 5.7$_{\textcolor{red}{-2.1}}$ & 3.5$_{\textcolor{red}{-4.3}}$ & 3.1  & 2.2$_{\textcolor{red}{-0.9}}$ & 7.7$_{\textcolor{blue}{+4.6}}$ & 3.7  & 2.1$_{\textcolor{red}{-1.6}}$ & 4.1$_{\textcolor{blue}{+0.4}}$ \\
InternVL$_{4B}$ & 61.2 & 61.1$_{\textcolor{red}{-0.1}}$ & 57.1$_{\textcolor{red}{-4.1}}$  & 8.3  & 15.9$_{\textcolor{blue}{+7.6}}$& 16.3$_{\textcolor{blue}{+8.0}}$ & 6.5  & 9.6$_{\textcolor{blue}{+3.1}}$ & 21.0$_{\textcolor{blue}{+15.0}}$ & 5.5  & 9.7$_{\textcolor{blue}{+4.2}}$ & 15.2$_{\textcolor{blue}{+9.7}}$ \\
InternVL$_{8B}$ & 33.2 & 64.9$_{\textcolor{blue}{+31.7}}$ & 51.5$_{\textcolor{blue}{+18.3}}$ & 9.9  & 13.4$_{\textcolor{blue}{+3.5}}$& 17.6$_{\textcolor{blue}{+7.7}}$ & 6.3  & 8.1$_{\textcolor{blue}{+1.8}}$ & 20.9$_{\textcolor{blue}{+14.6}}$  & 6.7  & 8.1$_{\textcolor{blue}{+1.4}}$ & 16.8$_{\textcolor{blue}{+10.1}}$ \\
InternVL$_{14B}$& 58.4 & 66.3$_{\textcolor{blue}{+7.9}}$ & 53.9$_{\textcolor{red}{-4.5}}$ & 15.3 & 23.2$_{\textcolor{blue}{+7.9}}$& 23.9$_{\textcolor{blue}{+8.6}}$ & 15.4 & 22.0$_{\textcolor{blue}{+6.6}}$ & 28.0$_{\textcolor{blue}{+12.6}}$ & 14.2 & 17.7$_{\textcolor{blue}{+3.5}}$ & 23.6$_{\textcolor{blue}{+9.4}}$ \\
InternVL$_{38B}$ & 48.7 & 68.1$_{\textcolor{blue}{+19.4}}$ & 56.1$_{\textcolor{blue}{+7.4}}$ & 24.5 & 28.0$_{\textcolor{blue}{+3.5}}$ & 26.1$_{\textcolor{blue}{+1.6}}$ & 27.2& 24.4$_{\textcolor{red}{-2.8}}$ & 29.9$_{\textcolor{blue}{+2.7}}$ & 23.9 & 23.1$_{\textcolor{red}{-0.8}}$ & 25.7$_{\textcolor{blue}{+1.8}}$\\
\hline
Claude & 68.8 & 74.8$_{\textcolor{blue}{+6.0}}$ & 66.9$_{\textcolor{red}{-0.9}}$ & 27.4 & 35.4$_{\textcolor{blue}{+6.0}}$ & 37.8$_{\textcolor{blue}{+8.4}}$ & 27.4 & 25.6$_{\textcolor{red}{-1.8}}$ & 36.6$_{\textcolor{blue}{+9.2}}$  & 28.9 & 29.7$_{\textcolor{blue}{+0.8}}$ & 35.6$_{\textcolor{blue}{+6.6}}$ \\
\bottomrule
\end{tabular}%
\end{table*}

\begin{figure}
    \centering
    \includegraphics[width=\linewidth]{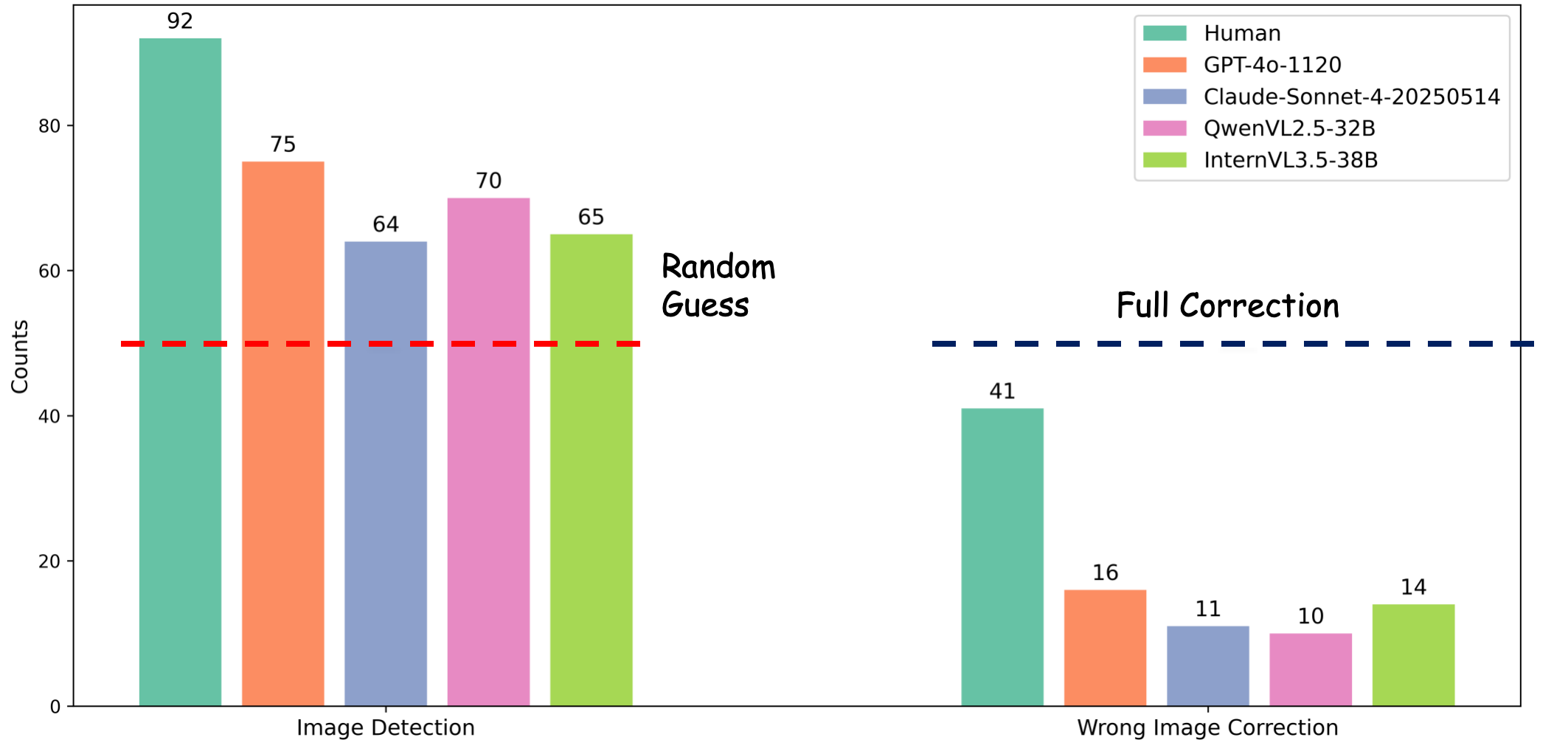}
    \caption{Performance of humans and VLMs on randomly selected images. Numbers indicate the counts of correctly classified images in image detection and fully corrected images in wrong image correction.}
    \label{fig:human_evaluation}
\end{figure}

\paragraph{\textit{How can humans detect and correct error tokens in an image?}}
We randomly select 100 images, with 50 in Chinese and 50 in English. For each language, 25 images contain errors and 25 are error-free. We then evaluate both open-source models (QwenVL-32B and InternVL-38B) and closed-source models (GPT-4o and Claude-Sonnet-4) under two criteria: (i) image-level detection (the ability to correctly classify whether an image contains errors) and (ii) wrong image correction (the ability to fully correct images containing errors). Two human annotators, one Chinese native speaker and one English native speaker, participate in the evaluation, and their responses are reviewed against the gold labels. Results in Figure~\ref{fig:human_evaluation} show that humans consistently outperform VLMs in error detection, with a success rate of 92\%. And the performance gap is even larger for error correction. For instance, humans successfully corrected 41 out of 50 erroneous images, while GPT-4o, the strongest model, managed only 16.

% \paragraph{Language Analysis} 
% The results categorized into Chinese and English are presented in Table \ref{tab:language_categorized_result_show}
% \input{table/language_categorized_result}

% \paragraph{How many characters are wrongly detected but correctly corrected}

\section{Solution Explorations}
\label{sec:solution_explorations}
\begin{table}[t]
\centering
\caption{Additional prompts for Joint OCR-Correction Paradigm and Background Information Enhancement, where the prompts are added to the direct prompt (Table~\ref{tab:prompt_visual_spelling}) as the first step.}
\label{tab:prompt_refinement}
\begin{tabular}{p{0.4\linewidth} p{0.4\linewidth}}
\hline
\multicolumn{2}{l}{\textbf{a) Joint OCR-Correction Paradigm}} \\
\hline
(Chinese) \textit{\chinese{先按照下面的格式提取并输出图片中的全部文本：“该图片中的文本为：......”}} & 
(English) \textit{Extract all text and then output it first in this format: "The extracted text from the image is:....."} \\
\hline
\multicolumn{2}{l}{\textbf{b) Background Information Enhancement}} \\
\hline
(Chinese) \chinese{请先对图片的内容（包括文本和环境）进行概括} & 
(English) \textit{Please first summarize the content of the image (including text and environment)} \\
\hline
\end{tabular}
\end{table}

% maybe the two method could be run in parallel
\subsection{Joint OCR-Correction Paradigm}
\label{subsec:joint ocr_correction}
According to the analysis in Section \ref{para:attention_visualization_analysis}, the end-to-end paradigm may lead to hallucinations about detected tokens. To mitigate this effect, we then formulate the task as a Unified Recognition-Correction Framework, where the model executes two tightly coupled stages within a single prompt: OCR followed by spelling correction. Numerous studies have applied OCR to enhance the performance of subsequent tasks~\cite{li-etal-2024-towards-real, ma-etal-2024-born, maheshwari2022benchmarkdatasetpostocrtext}. When the input is an image, the model must first extract all textual content and present it in a standardized format, followed by the prompt for error correction. The corresponding additional prompts could be found in Table \ref{tab:prompt_refinement} a).  By consolidating recognition and correction into a single, rule-governed framework, this approach establishes an efficient and consistent end-to-end pipeline for text correction. The results are shown in Table~\ref{tab:joint_ocr_f1_comparision}. 

We observe that introducing an OCR step before text correction leads to notable performance gains for most models. For example, the $F_{1}$ score of token-level correction (T-C) for QwenVL-32B improves from 36.5 to 44.0. Similarly, the performance of QwenVL-72B increases substantially, from 33.1 to 45.0, surpassing QwenVL-32B, and from 15.4 to 22.0 for InternVL-14B. These results highlight the effectiveness of incorporating OCR as an initial step to enhance error correction. We attribute this gain to the additional OCR stage, which improves the model's language understanding by providing explicit textual cues, thereby allowing the VLMs to leverage both the raw image input and the recognized text for more accurate correction. While for InternVL-38B, performance drops in T-D and T-C, while increasing in I-D and I-C. We also observe a performance drop in all evaluation tasks and levels for small models such as QwenVL-3B and InternVL-2B. This likely stems from their limited capacity to handle the extra noise introduced by imperfect OCR, making them less able to exploit the additional signal compared to larger models.

% At the same time, we also measured the model's performance under step by step settings. 

\subsection{Background Information Enhancement}
\label{subsec:joint_background_information}

\begin{figure*}
    \centering
    \includegraphics[width=\linewidth]{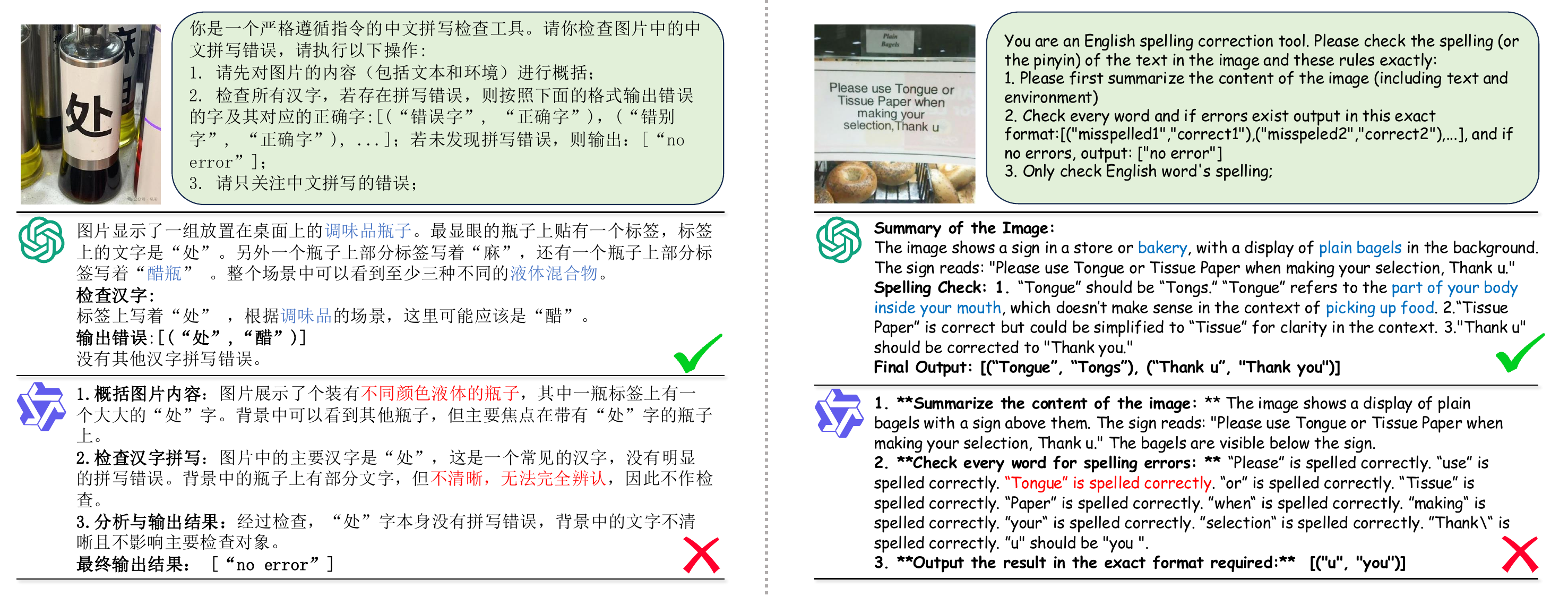}
    \caption{Examples of visual error correction by GPT-4o and QwenVL-32B under background information enhancement. The relevant background information is highlighted in \textcolor{blue}{blue}, while incorrect or vague descriptions are highlighted in \textcolor{red}{red}.
}
    \label{fig:bg_examples}
\end{figure*}
Based on previous solution exploration, we observe that the textual information alone is not complete, where the Joint OCR-correction paradigm only recognizes the explicit textual information and fails to exploit implicit background information, which could provide important conceptual cues. Moreover, errors in the OCR step can propagate downstream, further compounding errors in corrections. These observations raise a broader research question: \emph{Can VLMs effectively leverage background information for error correction?}

To investigate this, we propose a background information enhancement correction method, where the model is first asked to provide the background information of the image and then applies the correction based on this information. The additional prompts are listed in Table~\ref{tab:prompt_refinement} b). 

Our results in Table~\ref{tab:joint_ocr_f1_comparision} present the $F_1$ scores across four evaluation levels and reveal several key findings. First, for small models such as QwenVL-3B, injecting background information drops the performance of all tasks and levels, while for InternVL-2B, there's an increase in T-D (+4.6) and T-C (+0.4). Second, for larger models, both strategies consistently improve results. Notably, background information enhancement yields greater improvements for the InternVL family compared to OCR-based correction, likely because their native multimodal architecture is better equipped to integrate fine-grained visual cues with contextual semantics, allowing them to leverage background details more effectively for correction.

Case studies are displayed in Figure~\ref{fig:bg_examples}. The Chinese character \textcolor{red}{\chinese{处}} (\pinyin{chù}, place) on a sticker should be corrected to \textcolor{blue}{\chinese{醋}} (\pinyin{cù}, vinegar). The surrounding visual context-spice containers filled with black vinegar-provides a clear cue, yet none of the models generated the correct correction.  The phrase \textcolor{red}{tongues} should be corrected to \textcolor{blue}{tongs}, as the sign instructs on picking up food in a bakery. By comparing the examples of GPT-4o and QwenVL-32B, GPT-4o successfully obtained the precise and detailed background information from the picture, while QwenVL-32B only obtains general information. These examples suggest that background information can be highly beneficial, 
but only when models are capable of accurately identifying and leveraging it, 
which places higher demands on the reasoning ability of VLMs.

% Compared to the OCR-Correction paradigm, Background Information Enhancement provides more holistic information compared to text modality alone, providing a better solution for the corrections.

% \input{table/backgroud_correct_example}
\section{Conclusion}
\label{sec:conslusion}
This work introduced ReViCo, the first benchmark dedicated to evaluating vision-language models on real-world visual spelling correction in Chinese and English. By formulating tasks at both the image and token levels, ReViCo enables a systematic assessment of models' ability to detect and correct errors grounded in visual context. Our extensive experiments reveal that while current VLMs can partially identify erroneous tokens, they lag far behind human performance in producing correct corrections. We further explored two solution paradigms-Joint OCR-Correction and Background-Enhanced prompting-that substantially improve performance, particularly for larger models. These findings highlight the importance of combining explicit recognition with multimodal reasoning to overcome limitations of existing architectures. We hope ReViCo will serve as a foundation for advancing research on multimodal error correction and inspire the design of VLMs capable of handling fine-grained linguistic challenges in visually grounded scenarios.

{
    \small
    \bibliographystyle{ieeenat_fullname}
    \bibliography{main}
}
\end{document}